\documentclass[10pt,twocolumn,letterpaper]{article}

\usepackage{iccv}              % To produce the CAMERA-READY version
\usepackage{multirow} 
\usepackage{makecell}
\usepackage{amsmath, amssymb, algorithm, algorithmicx, algpseudocode}

\definecolor{iccvblue}{rgb}{0.21,0.49,0.74}
\usepackage[pagebackref,breaklinks,colorlinks,allcolors=iccvblue]{hyperref}

\def\paperID{13890} % *** Enter the Paper ID here
\def\confName{ICCV}
\def\confYear{2025}

\title{MixA-Q: Revisiting Activation Sparsity for Vision Transformers from a Mixed-Precision Quantization Perspective}

\author{
Weitian Wang$^{1,2}$, Rai Shubham$^1$, Cecilia De La Parra$^1$, Akash Kumar$^2$\thanks{Corresponding author}\\
$^1$Robert Bosch GmbH, Renningen, Germany, 
$^2$Ruhr University Bochum, Bochum, Germany\\
{\tt\small \{weitian.wang, shubham.rai, cecilia.delaparra\}@bosch.com, akash.kumar@ruhr-uni-bochum.de}
}
\begin{document}
% \baselinestretch{0.93}
\renewcommand{\baselinestretch}{0.94}

\maketitle
\definecolor{mygreen}{RGB}{0, 139, 0}
\definecolor{myred}{RGB}{179, 0, 0}
\definecolor{myredgreen}{RGB}{179, 139, 0}

\begin{abstract}
In this paper, we propose MixA-Q, a mixed-precision activation quantization framework that leverages intra-layer activation sparsity (a concept widely explored in activation pruning methods) for efficient inference of quantized window-based vision transformers. For a given uniform-bit quantization configuration, MixA-Q separates the batched window computations within Swin blocks and assigns a lower bit width to the activations of less important windows,
improving the trade-off between model performance and efficiency. We introduce a Two-Branch Swin Block that processes activations separately in high- and low-bit precision, enabling seamless integration of our method with most quantization-aware training (QAT) and post-training quantization (PTQ) methods, or with simple modifications. Our experimental evaluations over the COCO dataset demonstrate that MixA-Q achieves a training-free 
\textbf{1.35$\times$} computational speedup without accuracy loss in PTQ configuration. With QAT, MixA-Q achieves a lossless \textbf{1.25$\times$} speedup and a \textbf{1.53$\times$} speedup with only a 1\% mAP drop by incorporating activation pruning. Notably, by reducing the quantization error in important regions, our sparsity-aware quantization adaptation improves the mAP of the quantized W4A4 model (with both weights and activations in 4-bit precision) by 0.7\%, reducing quantization degradation by 24\%.
\end{abstract}
    
\section{Introduction}
\label{sec:intro}
% Recently, with the powerful representational capabilities of the self-attention mechanism, transformer-based model have shown dominance in many tasks including image classification\cite{}, object detection\cite{}, semantic segmentation\cite{}, etc., and thus being used as backbones in various applications. However, Vision Transformers (ViTs) require intensive computations, leading to excessive memory usage, high power consumption, and significant inference latency, which makes their deployment on resource-limited edge devices challenging.

Swin Transformer~\cite{liu2021swintransformerhierarchicalvision} is a representative vision transformer that generalizes well across different visual perception tasks, particularly dense prediction tasks such as object detection and segmentation~\cite{cheng2022maskedattentionmasktransformeruniversal}. While it improves upon the quadratic complexity of conventional ViT~\cite{dosovitskiy2021imageworth16x16words} by introducing a hierarchical structure and shifting window mechanisms, the self-attention operation remains computationally intensive.

With recent advancements in high-resolution imaging, 
% especially in the rapidly evolving automotive 
% industry, 
models process richer visual information, improving performance on dense tasks. However, this comes with increased computational overhead that results in latency bottlenecks for real-time applications like autonomous driving. Thus, reducing the computational cost associated with activations has become increasingly important. To tackle these computational challenges that arise from vision transformers and high-resolution inputs, activation pruning and quantization are two effective methods that improve efficiency with acceptable performance degradation.

Real-world images exhibit natural sparsity because not all pixels within an image are important. Activation pruning is a group of methods that exploits this activation sparsity to improve latency performance by selecting less important tokens~\cite{liu2023revisitingtokenpruningobject} or windows of transformers at runtime and skipping the computations associated with them. SparseViT~\cite{chen2023sparsevitrevisitingactivationsparsity} is a representative work for window-based ViTs like Swin Transformer and has achieved good latency improvement with negligible loss of accuracy. However, activation pruning methods have several limitations: 
\begin{itemize}
\item Require retraining to adapt the model for the pruned activations.
\item Significant accuracy degradation at high pruning ratios.
\item Performance is largely dependent on accurate window selections, which can't be guaranteed for \emph{Out-Of-Distribution} (OOD) inputs. This could limit their deployment on safety-critical applications.
\end{itemize}
% They require re-training to recover most of the performance and suffer from obvious performance degradation on high pruning ratios. Additionally, their performance is largely dependent on accurate window or token selection which can't be guaranteed for Out-Of-Distribution(OOD) inputs and adversarial inputs. This limits their deployment on safety-critical applications.

On the other hand, quantization is an effective and prevalent compression approach that reduces a model's computational cost by decreasing the representation precision of weights and activations. While most quantization methods use uniform-bit precision across the whole network, modern hardware increasingly supports the multi-bit precision, giving rise to mixed-precision quantization (MPQ) methods. The key insight in most prior MPQ methods is that different layers within the machine learning model don't have the same contribution to the final output and don't have the same robustness to quantization noise. However, they often overlook the intra-layer activation sparsity and focus on the inter-layer bit-width allocation. This offers another dimension to exploit activation sparsity from a mixed-precision quantization perspective: \emph{instead of skipping computations for less important regions, we can quantize them with lower bit width}.

\begin{figure}[t]
  \centering
  \begin{subfigure}[t]{0.48\linewidth}
    \includegraphics[width=\linewidth]{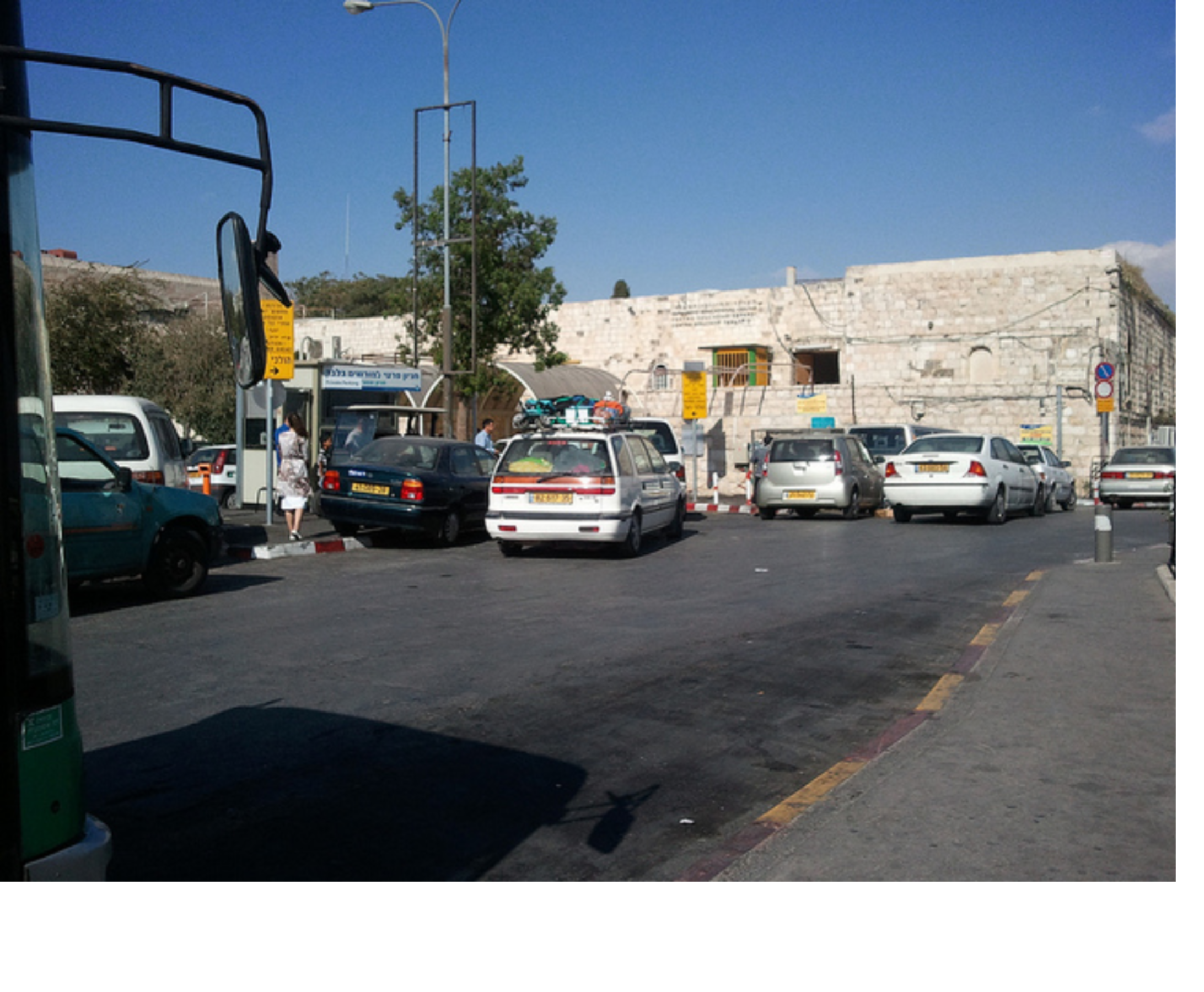}
    \caption{Example image from the COCO dataset~\cite{lin2015microsoftcococommonobjects}}
  \end{subfigure}
  \hfill
  \begin{subfigure}[t]{0.48\linewidth}
    \includegraphics[width=\linewidth]{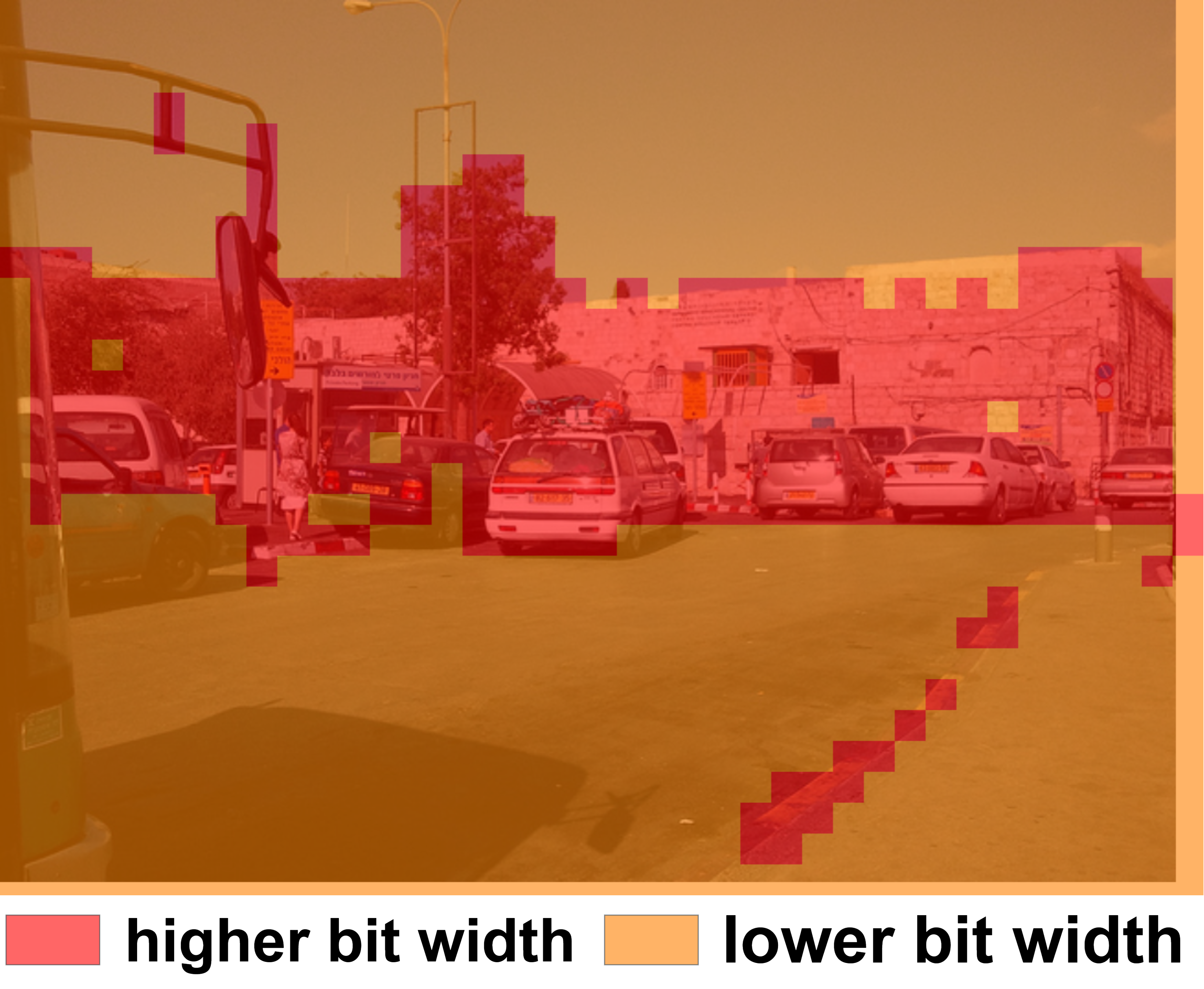}
    \caption{MixA-Q assigns lower bit width to the activation windows from less important regions to reduce computational cost.}
  \end{subfigure}
  \caption{Example for how MixA-Q leverages activation sparsity for efficient inference.}
  \label{fig:window-selection}
  \vspace{-2mm}
\end{figure}

% The key insight in most prior MPQ methods is that, different layers within the machine learning model don't have the same contribution to the final output and \textcolor{mygreen}{don't have the same robustness to quantization noise}. However, they overlook another important aspect: Unlike weights, which are randomly initialized and thus are usually equally important within the same layer. Activations naturally have spacial sparsity because not all pixels within an image is equally important. As mentioned before, for models with high-resolution input, the compression of activations has become more important. This leaves the mixed-precision quantization of activations an important but not yet investigated area.

% Activation pruning, where less important tokens or local windows of transformers are selected during runtime and their computations are skipped to improve latency, is a family of method which can improve the latency in floating point precision.

 In this paper, we bring the activation sparsity awareness from activation pruning to the quantization world and propose a novel mixed-precision activation quantization framework for Swin Transformer called MixA-Q. The key feature of Swin Transformer is that the feature map is separated into several local windows for batched multi-head attention~\cite{liu2021swintransformerhierarchicalvision}. 
\cref{fig:window-selection} illustrates how we leverage this feature of the Swin Transformer and the spatial sparsity of activations across local windows to reduce the computational cost. Using the L2 norm of the feature map, MixA-Q assigns high bit width to important foreground windows and keeps the rest in low bit width. We use an evolutionary search to optimize layer-wise compression ratios, accounting for varying computational costs and activation compression sensitivities. In the case of a high computation savings target, MixA-Q can incorporate activation pruning to achieve better performance than pure mixed-precision or pruning method. Our main contributions are summarized as follows:

% Different layers have different computational costs and sensitivities to the activation compression. We account for these variations by having a layer-wise compression ratio configuration that can be searched through evolutionary search. More details are discussed in \cref{sec:evol-search}.
 
 % Each stage of the Swin Transformer is assigned with different compression ratios that are defined as hyper-parameters. To find the best compression ratios for different stages, we use NSGA-II\cite{} to search for the ideal compression ratios that can best balance the accuracy and compression rate. In case of QAT where training is required, we use mixed-precision sparsity-aware adaptation to adapt the model for various compression ratios. Details are shown in Section XX.

\begin{enumerate}
    \item We propose a novel mixed-precision activation quantization framework for window-based ViTs that investigates the intra-layer activation sparsity for efficient inference of quantized models.
    \item We introduce a Two-Branch Swin Block architecture to replace the Swin blocks in Swin Transformer that enables mixed-precision execution of window-based attentions. With this architecture, MixA-Q can be integrated with most quantization-aware training (QAT) and post-training quantization (PTQ) methods with minor modification, making it flexible in terms of application.
    \item We adopt \emph{sparsity-aware quantization adaptation} (adapted from SpraseViT~\cite{chen2023sparsevitrevisitingactivationsparsity}) to adapt the model for mixed-precision activation sparsity. By improving the sampling algorithm with our \emph{uniform-sum compression ratio sampling}, we improve the sampling efficiency and reduce the fine-tuning time required.
    \item When integrated with QAT methods, we show that MixA-Q reduces the quantization error of important areas during the \emph{sparsity-aware quantization adaptation} through \emph{dynamic activation distillation}.
    \item Compared to directly applying activation pruning on the quantized model, our mixed-precision approach can be applied without training and is more robust to Out-Of-Distribution (OOD) inputs where the window selection is inaccurate.
\end{enumerate}

\noindent We evaluate MixA-Q on the COCO~\cite{lin2015microsoftcococommonobjects} dataset by integrating it with QAT and PTQ methods. MixA-Q achieves a lossless 1.35$\times$ computational speedup without training in PTQ configuration. On ultra-low bit width (4-bit), we integrate MixA-Q with the QAT method and achieve a lossless 1.25$\times$ speedup and a 1.53$\times$ speedup with only 1\% mAP drop by incorporating activation pruning. Additionally, through 
\emph{dynamic activation distillation} (detailed in \cref{sec:acc_enhance}), our sparsity-aware quantization adaptation improves the mAP of the quantized W4A4 model (weights and activations both in 4-bit precision) by 0.7\%, reducing quantization degradation by 24\%.

\section{Related Work}

\begin{figure*}[ht]
  \centering
  \begin{subfigure}[t]{0.57\linewidth}
    \includegraphics[width=\linewidth]{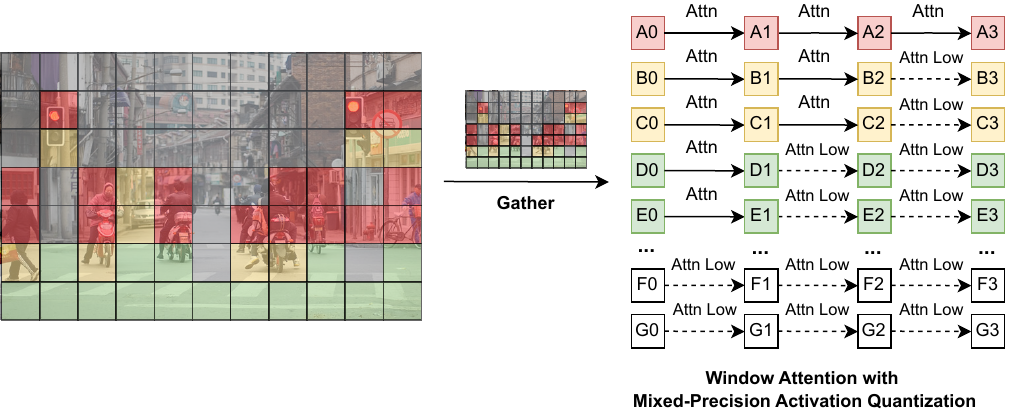}
    \caption{Our mixed-precision activation quantization attention block, which replaces the pruning block in SparseViT~\cite{chen2023sparsevitrevisitingactivationsparsity}. Instead of skipping the computation of less important patches and copying the features from the last computation like \cref{fig:2b}, we keep the computations in low precision.}
    \label{fig:2a}
  \end{subfigure}
  \hfill
  \begin{subfigure}[t]{0.33\linewidth}
    \includegraphics[width=\linewidth]{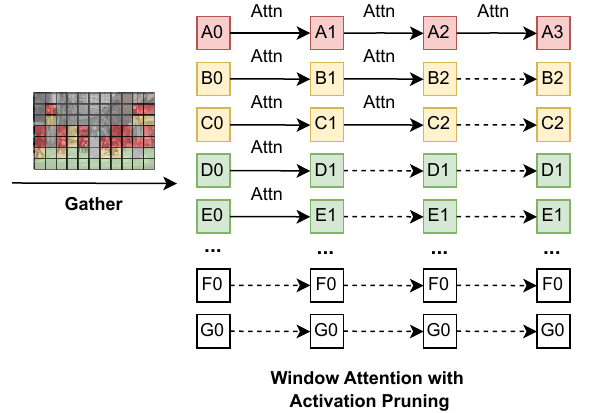}
    \caption{The window attention with activation pruning from SparseViT~\cite{chen2023sparsevitrevisitingactivationsparsity}. The computations on less important windows are skipped, and the latest features are forwarded.}
    \label{fig:2b}
  \end{subfigure}
  \caption{Overview of how our MixA-Q replaces the window pruning in SparseViT~\cite{chen2023sparsevitrevisitingactivationsparsity} to mixed-precision execution. Note that the LN and FFN following the window attention are omitted here.}
  \hfill
  \label{fig:short}
\end{figure*}

\subsection{Vision Transformers}
Vision Transformers (ViTs) adapt the self-attention mechanism from NLP \cite{vaswani2023attentionneed} to image processing~\cite{dosovitskiy2021imageworth16x16words}. While ViT showed impressive performance on image classification, it faced challenges with dense prediction tasks like object detection. Swin Transformer~\cite{liu2021swintransformerhierarchicalvision} introduces a hierarchical structure and shifted windowing to efficiently handle high-resolution images. Its linear computational complexity to image size and strong performance on dense tasks like detection and segmentation have made it particularly effective for a wide range of computer vision applications and has become the backbone for many vision models~\cite{cheng2022maskedattentionmasktransformeruniversal, zhang2021knetunifiedimagesegmentation}.
\subsection{Model Quantization}
Model quantization aims to reduce the computational and memory requirements of machine learning models by lowering the precision of weights and activations. By retraining the network and allowing the model to compensate for the quantization error, QAT methods can achieve very competitive quantization performance. For the image classification task, state-of-the-art QAT methods~\cite{liu2023oscillationfreequantizationlowbitvision,li2022qvitaccuratefullyquantized} have achieved higher accuracy than the original floating-point model in ultra-low-bit precision (4 bits). Due to the high resource and time usage required by QAT methods, PTQ methods are favored for dense prediction tasks~\cite{liu2024pq, li2023repqvitscalereparameterizationposttraining} and LLMs~\cite{liu2024qllmaccurateefficientlowbitwidth}. Many prior PTQ methods~\cite{nagel2020downadaptiveroundingposttraining,nagel2019datafreequantizationweightequalization} have achieved impressive results on CNNs.
% yet have poor performance on ViTs. 
Recently, many works have also investigated PTQ methods for ViTs in image classification and object detection tasks, such as RepQ~\cite{li2023repqvitscalereparameterizationposttraining} and EQR~\cite{zhong2025accurateposttrainingquantizationvision}.

\subsection{Mixed-precision Quantization}
Unlike uniform-bit quantization that aims to reduce the quantization error, mixed-precision quantization (MPQ) aims to achieve better precision and performance trade-off by assigning different bit widths to activations and/or weights. Existing mixed-precision quantization methods can be classified into search-based approaches~\cite{wang2020apqjointsearchnetwork} and criterion-based approaches~\cite{dong2019hawqhessianawarequantization,dong2019hawqv2hessianawaretraceweighted,sun2022entropy}. Most of these works focus on the mixed-precision quantization of weights, with some~\cite{dong2023emqevolvingtrainingfreeproxies,ma2022ompqorthogonalmixedprecision} still using uniform bits for activation. For the mixed-precision quantization of activations, Kim \etal~\cite{Kim_2024} found that the activation instability due to bit selection can affect the performance largely and propose to train a meta-network for the bit selection of activations for different layers. PMQ\cite{xiao2023patchwisemixedprecisionquantizationvision} saves the computation of MLPs following attention layers by adapting the bit allocation of tokens according to the attention scores. Granular-DQ\cite{wang2024thinkinggranularitydynamicquantization} dynamically quantizes different patches based on their information density in super-resolution tasks.

\subsection{Activation Pruning}
Pruning the computation on less important areas can improve latency and memory consumption without apparent accuracy loss. Due to the sparse nature of the attention mechanism, token pruning methods for ViTs have been well investigated, including using gating networks~\cite{kong2022spvitenablingfastervision,rao2021dynamicvitefficientvisiontransformers}, attention scores~\cite {fayyaz2022adaptivetokensamplingefficient,liang2022patchesneedexpeditingvision}, etc. SparseViT~\cite{chen2023sparsevitrevisitingactivationsparsity} prunes local windows of Swin Transformer on dense prediction tasks and achieves competitive performance. In activation pruning, the selected windows or tokens considered less important are \textit{pruned}, meaning that the computations over these windows or tokens are skipped for certain layers, and the features from previous layers are simply forwarded. 
\section{Mixed-Precision Activation Quantization}

In this section, we present the overall MixA-Q framework.  Within the framework, we explore the adaptation of activation pruning to mixed-precision activation quantization. Instead of pruning the less important windows or tokens, we process them with the low-precision branch, thus saving the computation complexity without losing too much information. 

\subsection{From Pruning to Mixed-Precision}
\label{sec:from_prunning_to_mixed_precision}
% The detailed advantages over doing this are illustrated in \cref{sec:mpaq advantage}.
Before going into the details of our framework, we first answer this question: 

\textit{Why do we need mixed-precision activation quantization if we can directly apply the activation pruning method on the quantized model to further optimize it?} 

It is because instead of skipping the computation of less important regions, retaining them in lower precision has several advantages:
\begin{itemize}
    \item Mixed-precision activation quantization doesn't change the computation flow of the network and thus can be integrated with PTQ methods without training.
    \item At high pruning ratios, most activation pruning methods suffer from obvious accuracy degradation due to drastic information loss. In this case, retaining some less important but relevant windows at low precision is beneficial. More details are discussed in \cref{sec:ap-incorp} and \cref{sec:pareto}.
    \item The robustness of activation pruning methods relies on simple window/token selection criteria~\cite{chen2023sparsevitrevisitingactivationsparsity} or light-weight layers~\cite{liu2023revisitingtokenpruningobject}. In the case of out-of-distribution inputs, \eg street in foggy weather, where window/token selection is inaccurate, activation pruning methods may suffer from drastic performance degradation due to pruning important patches. On the other hand, the performance of MixA-Q in the worst case is bounded below by the low-precision branch. We will further investigate this claim in \cref{sec:robustness}. 
    \item By updating the model only with important activations, MixA-Q can reduce the model's overall quantization error other than the latency improvement through \emph{dynamic activation distillation} (more details are in \cref{sec:acc_enhance}). 
\end{itemize}

\noindent Hence, we adapt the window-level activation pruning proposed in SparseViT~\cite{chen2023sparsevitrevisitingactivationsparsity} to a mixed-precision activation quantization framework for Swin Transformers~\cite{liu2021swintransformerhierarchicalvision}. 

% Token pruning methods are versatile for wide range of ViT variants. However, the computation flow of different tokens are not independent, and hence putting them into different precisions makes different columns of a matrix during multiply–accumulate (MAC) operations to be in different precisions. Such computations often come with deployment overheads and are heavily dependent upon the hardware platform. 

\paragraph{Replace Pruning with MP Execution} 

Swin Transformer separates the feature map into \textbf{non-overlapping} image windows (\eg 7×7)  and performs multi-head self-attention (MHA) upon them to extract local features. Each MHA is followed by a layer normalization (LN) and a feed-forward network (FFN) layer. Taking advantage of this window-based attention design of Swin Transformer, SparseViT first gathers the features from the windows with the highest importance scores, then runs computations on the selected windows, and finally scatters them back to the original feature map. As shown in \cref{fig:2b}, after certain windows are pruned at some point, the features stop being further processed, \eg ``D1'' is only forwarded after being pruned at the second block. 
 
Instead of skipping the computations on less important windows, MixA-Q compresses them by assigning a lower bit width to them (annotated as ``Attn Low'' in \cref{fig:2a}). In this way, the network's original computations upon those windows are still kept, but their computational cost is reduced.

% \paragraph{Window Selection and Execution} As in SparseViT\cite{chen2023sparsevitrevisitingactivationsparsity},
% the importance of each window is defined its $L_2$ activation magnitude. Given a compression ratio, we gather those windows with the highest importance scores and feed them to the high-precision branch (detailed in \cref{sec:ts-sb}) and then feed the rest of the windows to the low-precision branch. Finally, the output features of the two branches are scattered back to a full feature map. For each stage the importance score is only computed once before the execution of the first swin block and then shared across the whole stage.
\subsection{Two-Branch Swin Block}
\begin{figure*}[t]
    \centering
    \includegraphics[scale=0.7, width=0.9\linewidth]{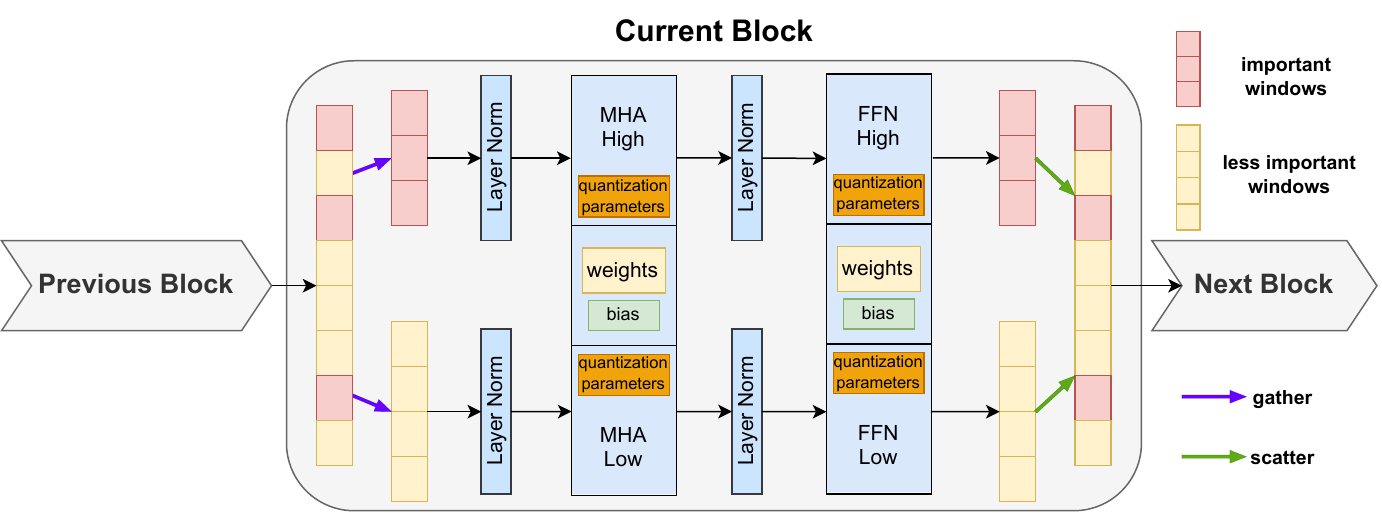}
    \caption{Two-branch swin block that replaces the original swin blocks in Swin Transformer. The windows are gathered into high- and low-precision windows according to the importance score. They go through the high- and low-precision branches separately and then are scattered back to the feature map.}
    \label{fig:ts-sb}
\end{figure*}
\label{sec:ts-sb}
To support the separated computation on high- and low-precision windows, we introduce the Two-Branch Swin Block in \cref{fig:ts-sb} that replaces the original swin blocks within Swin Transformer. Each two-branch swin block takes as input a sequence of activation windows. As in SparseViT, the importance score of each window is defined as its L2 norm. For each stage, the importance score is only computed once before the execution of the first swin block and then shared across the whole stage. We call the layers that perform computation on high-precision windows \textit{high-precision branch} and the layers that perform computation on low-precision windows \textit{low-precision branch}. Given a compression ratio (detailed in \cref{sec:evol-search}), we gather those windows with the highest importance scores, feed them to the high-precision branch, and then feed the rest of the windows to the low-precision branch.

The layer normalizations are duplicated for the high- and low-branches to address the distribution shift after the separation. After the duplication, parameters within layer normalizations can be updated independently during fine-tuning (in QAT) or reparameterization (in PTQ). On the other hand, multi-head attention layers (MHA) and feed-forward networks (FFN) are created as \textit{shadow layers} that share the same underlying weights and bias while independent step sizes and zero points can be kept for the high- and low-precision branches. During training, the gradients of shadow layers from high- and low-precision branches will flow through the same set of parameters and be accumulated by the autograd engine of Pytorch.

\subsection{Evolutionary Search of Compression Ratios}
\label{sec:evol-search}
Following the design of SparseViT~\cite{chen2023sparsevitrevisitingactivationsparsity}, each two consecutive swin blocks (one with shifted windows and one without) are assigned a fixed compression ratio, which is the proportion of windows that should be processed by the low-precision branch. The compression ratios are hyperparameters that can be optimized with an evolutionary search (detailed in \cref{sec:evol-search}). 
The search for optimal compression ratios can be formed as a bi-objective optimization problem:
\begin{itemize}
    \item Objectives: computational savings, mAP (mean average precision)
    \item Variables: compression ratios for each stage
\end{itemize}

The computational saving is computed as the proportion of bit operations that are saved by applying the compression ratios. mAP is the mean average precision on the COCO val dataset. In the case of Swin-Tiny, there are 6 compression ratios for the 12 swin blocks in the model. To narrow the search space, we discretize the compression ratios to \( \{0\%, 10\%, ...,80\%\} \). We use the NSGA-II~\cite{deb2002fast} algorithm implemented in pymoo~\cite{9078759} to search for the optimal ratios iteratively.
\vspace{-2mm}
\paragraph{Sparsity-aware Quantization Adaptation}
For PTQ methods, the same quantized model is used for evaluation during the evolutionary search. However, for QAT methods, the model needs to be fine-tuned after changing the compression ratios, which makes searching compression ratios inapplicable. Thus, we adopt a sparsity-aware quantization adaptation (SAQA), which is similar to the sparsity-aware adaptation (SAA) introduced in SparseViT~\cite{chen2023sparsevitrevisitingactivationsparsity}. At first, we quantize the model with uniform-bit QAT method on the high precision. Then, we perform the SAQA by training the model while randomly sampling a set of compression ratios at each iteration, i.e., a random proportion of activations is assigned to the low-precision branch. After SAQA, the model achieves a good estimated performance for different compression ratio configurations without re-training.

\vspace{-3mm}
\paragraph{Uniform-sum Compression Ratio Sampling}
To make evolutionary search and SAQA more sample efficient, we propose an algorithm called \emph{uniform-sum compression ratio sampling}. SparseViT uses a naive sampling algorithm, where the ratios for each layer are uniformly sampled from \( \{0\%, 10\%, ...,80\%\} \). This causes the averaged pruning ratio of the model to have a scaled Irwin–Hall distribution (uniform sum distribution), a bell-shaped distribution centered at 40\%. This makes the sampling of evolutionary search inefficient. Furthermore, the adapted model shows inaccurate estimations on compression ratios that have an average far away from the center. In our \emph{uniform-sum ratio sampling}, we begin by randomly sampling a scalar value \( S \) from a predefined range, representing the target sum of compression ratios across all blocks. Given \( S \), we then sample the individual compression ratios \( \{r_0, r_1, \dots, r_n\} \) using a Dirichlet distribution, ensuring that their sum is constrained to \( S \). To enforce an upper bound on each ratio, we apply a rejection sampling criterion: if any \( r_i \) exceeds \( 0.8 \), the sampled set is discarded, and the process is repeated until a valid set is obtained. More details can be found in Appendix 6.1.

\subsection{Activation Pruning Incorporation}
\label{sec:ap-incorp}
Activation pruning and mixed-precision quantization are not mutually exclusive. Compared to mixed-precision activation quantization, pruning-based methods have higher efficiency in terms of activation sparsity exploitation at the cost of more information loss. For each less important window, the pruning-based method saves 100\% of its computations by skipping them. On the other hand, mixed-precision activation quantization saves 50\% of computations by downgrading the precision to half of the bit widths. In the case of a high computation savings target, pruning the most unimportant windows while keeping the less important but still relevant windows in lower precision can leverage the activation sparsity efficiently while avoiding drastic information loss due to aggressive pruning.

With this insight, we also incorporate activation pruning in MixA-Q that can be activated when integrating with QAT methods. When activation pruning is adopted, each two consecutive swin blocks are assigned an additional pruning ratio, apart from the compression ratio. In the window selection phase, the windows with the lowest importance scores will be pruned, and the computations upon them will be skipped as in \cref{fig:2b}. Then, MixA-Q continues separating the rest of the windows and processing them in two precisions. 
\section{Experiments}
In this section, we evaluate MixA-Q by integrating it with QAT and PTQ methods respectively. We evaluate our method's computation savings with relative computation cost or speedup in terms of bit operations (BOPs) of the backbone, as our method is only applied to the backbone.

\subsection{QAT Result on COCO}
\label{sec:qat_coco}
% In this section, we show our method's evaluation result on QAT and ultra-low bit widths: mixed-precision of 4 bits and 2 bits.
\paragraph{Setup} 
We integrate our method with a QAT method, OFQ~\cite{liu2023oscillationfreequantizationlowbitvision}, to quantize a Mask R-CNN~\cite{he2018maskrcnn} with Swin-Tiny~\cite{liu2021swintransformerhierarchicalvision} backbone. The COCO~\cite{lin2015microsoftcococommonobjects} dataset is used for training and evaluation. We first quantize the floating-point model with OFQ on W4A4 (weight and activations both in 4-bit precision). This W4A4 OFQ model serves as a baseline model for evaluating our mixed-precision activation quantization method, MixA-Q, and the activation pruning method, SparseViT~\cite{chen2023sparsevitrevisitingactivationsparsity}. When applying MixA-Q, the less important windows are assigned a bit width of 2.

For all the metrics of SparseViT reported in this section on the W4A4 OFQ model, we follow the procedure reported in the SparseViT~\cite{chen2023sparsevitrevisitingactivationsparsity} paper. We first train the sparsity-aware adapted model and use evolutionary search to search for the optimal pruning ratios, and finally fine-tune the model. The optimal pruning ratios that we found on the W4A4 OFQ model and the performance degradation after applying the pruning align with the metrics reported in~\cite{chen2023sparsevitrevisitingactivationsparsity}.

\subsubsection{Analyses and Comparisons}
\label{sec:pareto}
\begin{figure}[t]
    \centering
    \includegraphics[scale=0.5, width=0.95\linewidth]{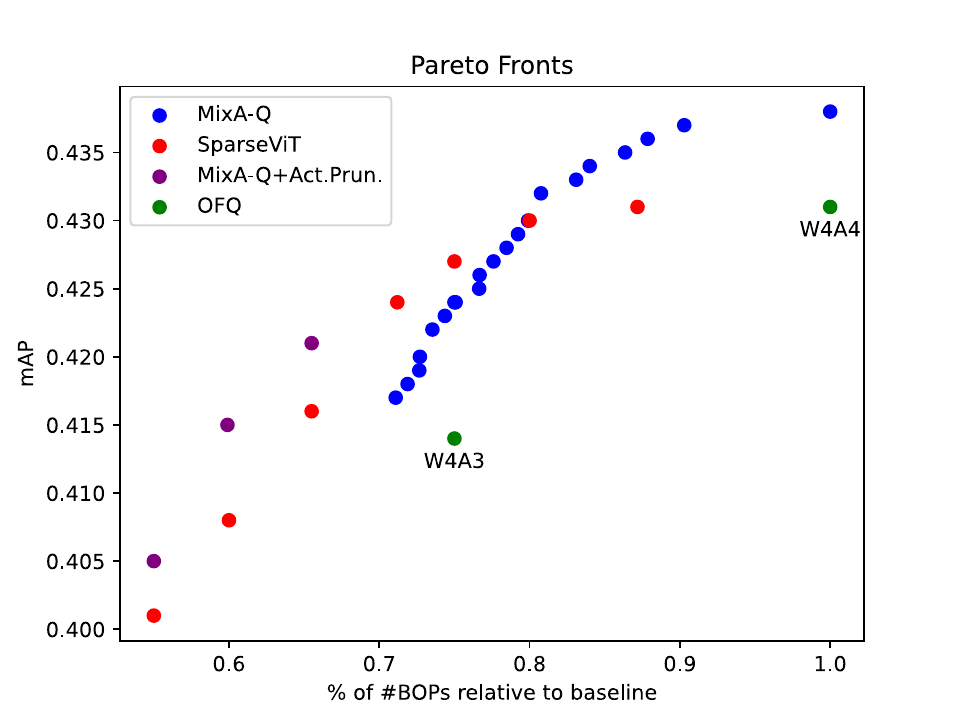}
    \caption{Pareto fronts of different methods. The x-axis is the relative computation cost in terms of bit operations (BOPs) to the baseline model (W4A4 OFQ model). ``MixA-Q+Act.Prun.'' represents MixA-Q incorporated with 30\% activation pruning.}
    \label{fig:pareto-qat}
\vspace{-3mm}
\end{figure}

% Prior to the quantitative analyses, 
First, we present the Pareto fronts of different methods in \cref{fig:pareto-qat} to offer insight into their performance variations at different relative computational costs (abbreviated as relative cost) in terms of bit operations (BOPs). It can be observed from the figure that the precision and performance trade-off of MixA-Q and SparseViT is better than the uniform-bit OFQ. At higher relative costs ($\geq0.8$), MixA-Q outperforms other works in terms of mAP. Notably, MixA-Q shows higher mAP than the OFQ model in W4A4. The reason behind this will be further investigated in \cref{sec:acc_enhance}. However, at relative costs $\leq$ 0.75, we can observe that SparseViT starts to have higher mAP than MixA-Q due to its higher efficiency in terms of sparsity utilization, as discussed in \cref{sec:ap-incorp}. To leverage this advantage of activation pruning while avoiding the drastic information loss, we incorporate MixA-Q with 30\% overall activation pruning. As shown by the purple dots in \cref{fig:pareto-qat}, the MixA-Q incorporated with activation pruning outperforms the pure activation pruning method, SparseViT, at relative costs ($\leq0.7$). 
% lower than 0.7.

We also report numerical results in \cref{tab:qat} for high relative costs ($\geq$ 0.75) and \cref{tab:mix} for low relative costs ($\leq$ 0.7). As shown in \cref{tab:qat}, MixA-Q achieves 1.24$\times$ speed up with slightly better mAP than the W4A4 OFQ model. Compared with the W4A3 OFQ model where precision of activations are uniformly downgraded to 3 bits, we achieve a higher 1.35× speed up with less than 1\% mAP loss (1.7\% for W4A3 OFQ). From \cref{tab:mix}, it can be noticed that by incorporating activation pruning, MixA-Q achieves a 1.53$\times$ speedup with only 1\% mAP loss. For even higher speedups of 1.66$\times$ and 1.82$\times$, MixA-Q with activation pruning continuously achieves higher mAP than SparseViT at the same relative costs. This shows that retaining less important but still relevant windows in lower precision helps recover some performance degradation caused by aggressive pruning and can achieve better mAP at a comparable relative cost.

\begin{table}[t]
\centering
\begin{tabular}{l|llll}
\hline
Method           & Act Bit     & \#BOPs (T)  & mAP        \\ \hline
Full-Precision   & 32           & 88.42  & 46.0     \\ \hline
\multirow{2}{*}{OFQ~\cite{liu2023oscillationfreequantizationlowbitvision}} 
                & 4           & 11.05  & 43.1    \\
                & 3           & 8.29 (1.33$\times$)  & 41.4    \\ 
\hline
\multirow{2}{*}{SparseViT~\cite{chen2023sparsevitrevisitingactivationsparsity}} 
                & 3.2*     & 8.92 (1.24$\times$)  & 43.1    \\
                & 2.96*     & 8.28 (1.33$\times$)  & 42.7    \\ \hline
\multirow{3}{*}{MixA-Q (Ours)} 
                & 4        & 11.05 & \textbf{43.8}    \\
                & 3.2*    & 8.92 (1.24$\times$)  & \textbf{43.2}    \\
                & 2.94*   & 8.21 (1.35$\times$)  & 42.3    \\
\hline
\end{tabular}
\caption{Quantization results of different methods on the COCO dataset. mAP is the bounding box average precision for object detection. Act Bit represents the activations' equivalent bit width. * indicates that the bit width is a computationally weighted average. \#BOPs (T) is the number of bit operations in trillion scale. The scale shown after \#BOPs indicates the computational speedup compared to the W4A4 OFQ model. MixA-Q and SparseViT are both applied to the model quantized with OFQ in W4A4 precision.}
\label{tab:qat}
\end{table}

\begin{table}[t]
\centering
\begin{tabular}{l|llll}
\hline
Method           & Act Bit     & \#BOPs (T)  & mAP        \\ \hline
OFQ~\cite{liu2023oscillationfreequantizationlowbitvision}   & 4           & 11.05  & 43.1      \\ \hline
\multirow{4}{*}{SparseViT~\cite{chen2023sparsevitrevisitingactivationsparsity}} 
                & 2.6*     & 7.24 (1.53$\times$)  & 41.4    \\
                & 2.4*     & 6.63 (1.66$\times$)  & 40.8    \\
                & 2.2*     & 6.08 (1.82$\times$)  & 40.1    \\ \hline
\multirow{3}{*}{\makecell{MixA-Q\\+Act.Prun.}} 
                & 2.6*     & 7.24 (1.53$\times$) & \textbf{42.1}    \\
                & 2.4*     & 6.63 (1.66$\times$)  & \textbf{41.5}    \\
                & 2.2*     & 6.08 (1.82$\times$)  & \textbf{40.5}    \\
\hline
\end{tabular}
\caption{Results of SparseViT and MixA-Q with activation pruning (denoted as MixA-Q+Act.Prun.) at high speedups.}
\label{tab:mix}
\vspace{-2mm}
\end{table}

\subsubsection{Dynamic Activation Distillation}
\label{sec:acc_enhance}
After applying SAQA on the W4A4 OFQ model, our MixA-Q model shows a mAP enhancement as observed in \cref{fig:pareto-qat} and \cref{tab:qat}. In this section, we explore the reason behind it. We confirm that this is a benefit brought by our SAQA because the mAP enhancement can't be observed by continuing training the W4A4 OFQ model or applying the pruning-based SAA from SparseViT.

From \cref{tab:saa-sqnr}, we can observe that after applying SAQA, the model has a higher mean \emph{signal-to-quantization-noise ratio} (mSQNR$\uparrow$) at later stages but lower mSQNR at early stages. We attribute this phenomenon to the fact that during SAQA, only gradients from important windows can flow through the high-precision (4-bit) branch. This process guides the model to focus on reducing the quantization error of important windows during training. We call this process \emph{dynamic activation distillation} as it selects important parts from activations to better update the quantization-related parameters.

\begin{figure*}[htbp]
\centering
\includegraphics[width=\linewidth]{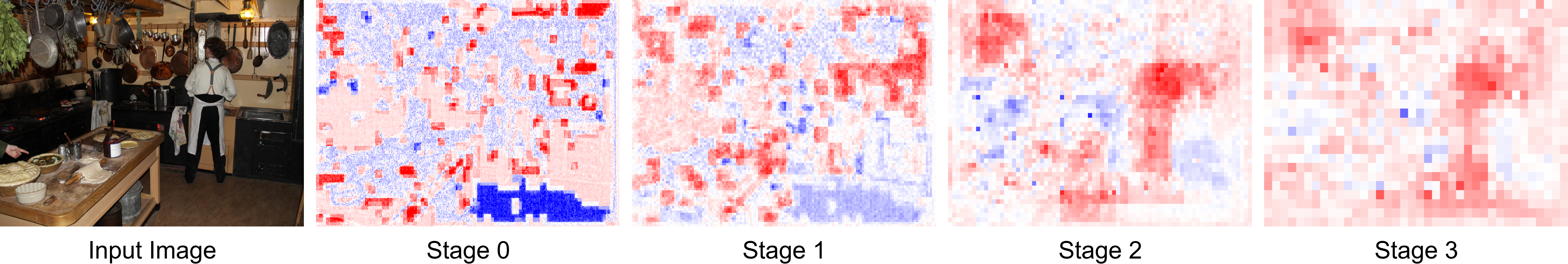}
\caption{Signal-to-quantization-noise ratio (SQNR$\uparrow$) changes across different stages before and after SAQA. Red implies higher SQNR after SAQA, and blue implies lower SQNR. At early stages, the model learns to decrease the quantization noise of important areas, which leads to an increase in overall SQNR at later stages.}
\label{fig:sqnr_change}
\end{figure*}

In \cref{fig:sqnr_change}, we visualize the SQNR changes after applying SAQA, which aligns with our explanation and the results in \cref{tab:saa-sqnr}. At early stages (Stage 0 and Stage 1), quantization noise is reduced (red windows) for foreground windows but increases (blue windows) for background windows, leading to lower overall SQNR due to the abundance of background windows. At stage 2 and 3, the quantization noise of more windows is reduced because the important windows from early stages contribute more to the high-level features at later stages, corresponding to the mSQNR increase at later stages in \cref{tab:saa-sqnr}. In general, SAQA guides the model to reduce quantization noise of important windows and thus improve the mAP.

\begin{table}[htbp]
\centering
\begin{tabular}{c|cc}
\hline
\multirow{2}{*}{} & \multicolumn{2}{c}{mean SQNR$\uparrow$} \\
                  & before SAQA    & after SAQA   \\ \hline
Stage 0           & 13.70          & 13.60        \\
Stage 1           & 9.70           & 9.50         \\
Stage 2           & 6.90           & 6.96         \\
Stage 3           & 7.81           & 8.14         \\ \hline
\end{tabular}
\caption{Mean SQNR at different stages before and after the SAQA over 1000 samples. The SQNR is computed between the different stages' output features of the quantized model and those of the floating-point model.}
\label{tab:saa-sqnr}
\vspace{-2mm}
\end{table}

% \begin{table}[htbp]
% \centering
% \begin{tabular}{l|ll}
% \hline
% MeanSQNR $\uparrow$ & before SAA & after SAA \\ \hline
% Stage 0 high                    & 1.07          & 1.17          \\
% Stage 1                         & 0.443          & 0.436          \\ \hline
% Stage 2                         & 0.332          & 0.331          \\
% Stage 3                         & 0.121          & 0.119          \\ \hline
% \end{tabular}
% \caption{Mean accumulative quantization error at different stages of the W4A4 OFQ model before and after the mixed-precision activation SAA over 100 random samples.}
% \end{table}
\subsubsection{Robustness of Pure Activation Pruning}
\label{sec:robustness}
Performance on the test dataset is not the only criterion for evaluating the quality of a method. The ability to detect objects regardless of image distortions or weather conditions is crucial for safety-critical systems like autonomous driving. SparseViT relies on the L2 norm of features to effectively decide the windows' importance. However, for the first stage, the feature map is only dependent on the local patch embedding layer, which lacks the global view and can be largely affected by natural image shift. Applying activation pruning with inaccurate window selection will result in the pruning of important parts, which leads to catastrophic performance degradation. On the other hand, although our MixA-Q is also affected negatively by inaccurate window selection, the computations of the wrongly compressed windows are not skipped but kept in low precision. In the worst case, where only important windows are computed with low precision, the model's performance is still guaranteed by the low precision branch. 

We use the weather scenario from the COCO-O~\cite{mao2023cocoobenchmarkobjectdetectors} dataset for evaluating the model's robustness on out-of-distribution inputs, which consists of real-world images that contain objects in challenging weather conditions, \eg rain, snow, and fog.

\begin{table}[htbp]
\centering
\begin{tabular}{l@{\hskip 6pt}|c@{\hskip 6pt}c@{\hskip 6pt}c@{\hskip 6pt}c@{\hskip 6pt}c}
\hline
                        & mAP         & mAP$_o$ &    $r_d^o \downarrow$ & mAP$_w$  & $r_d^w\downarrow$  \\ \hline
OFQ, W4A4               & 43.1        &  34.2   &    20.6               &  /       &        /              \\
MixA-Q,1.4$\times$      & 41.5        &  30.5   &    26.5               &  28.8  &      30.6                 \\
SparseViT, 1.4$\times$  & 42.4        &  29.6   &    30.2               &  24.0  &      43.4                 \\ \hline
\end{tabular}
\caption{mAP on COCO~\cite{lin2015microsoftcococommonobjects}, COCO-O~\cite{mao2023cocoobenchmarkobjectdetectors} (mAP$_o$) dataset, and on COCO with worst window selection (mAP$_w$). The worst window selection is achieved by calculating the importance score in reverse order. 1.4$\times$ means the model is compressed by MixA-Q or pruned by SparseViT to reach the speedup of 1.4$\times$ in terms of computations. $r_d^o$ and $r_d^w$ is the degradation rate of mAP$_o$ and mAP$_w$ with respect to the mAP.}
\label{tab:robust}
\vspace{-3mm}
\end{table}

\begin{figure}[htbp]
  \centering
  \begin{subfigure}[t]{0.48\linewidth}
    \includegraphics[width=\linewidth]{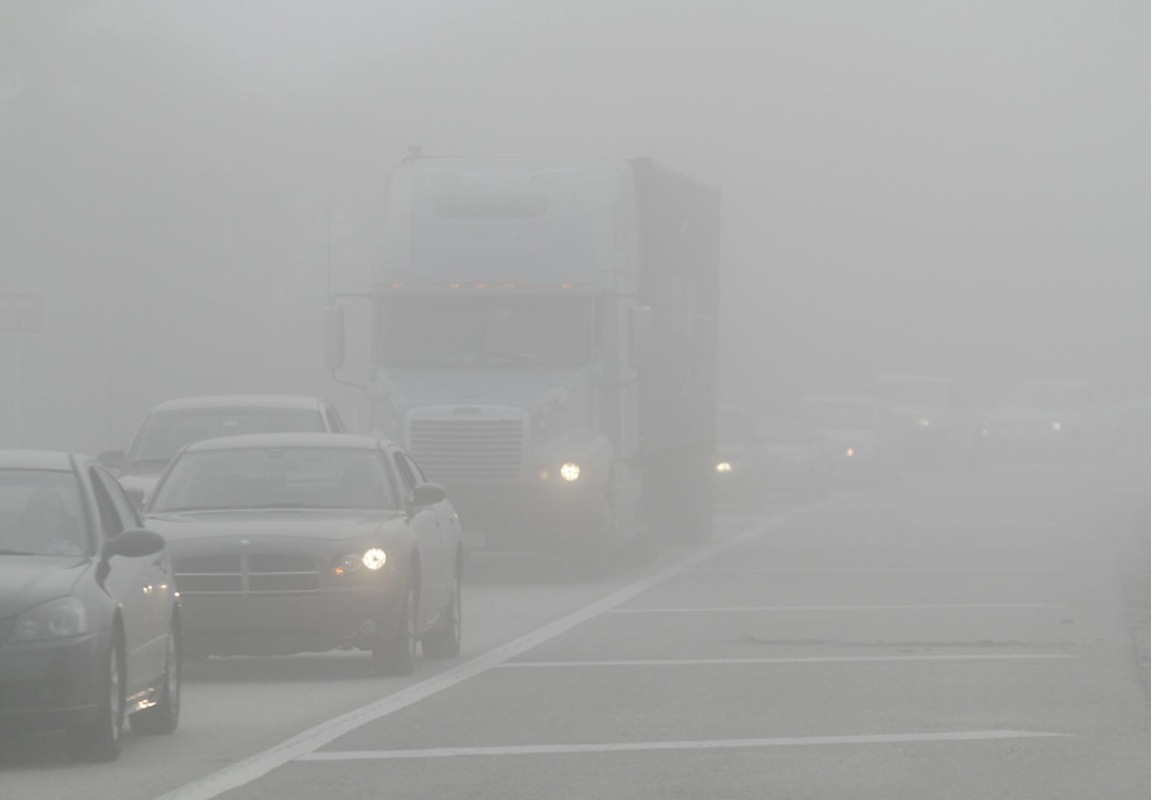}
    \caption{Foggy weather}
  \end{subfigure}
  \hfill
  \begin{subfigure}[t]{0.48\linewidth}
    \includegraphics[width=\linewidth]{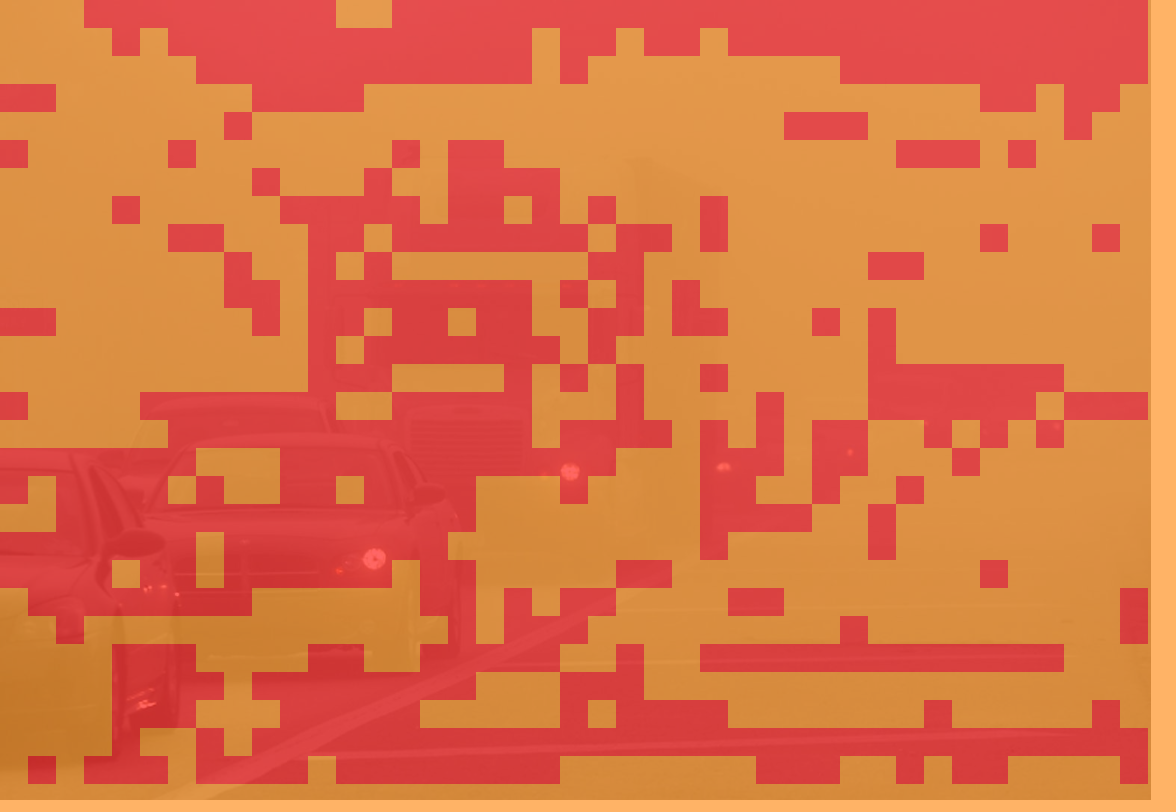}
    \caption{Window selection of SparseViT at stage 0. Red windows are important windows selected to be kept and orange windows are pruned.}
  \end{subfigure}
  \caption{Visualization of the window selection of SparseViT over an OOD sample from COCO-O}
  \label{fig:sparse_win_selection}
  \vspace{-2mm}
\end{figure}

From \cref{tab:robust}, it can be observed that although SparseViT and MixA-Q both cause a higher decay rate on OOD inputs than the uniform-bit quantized model, but the model pruned by SparseViT has a worse degradation rate than MixA-Q. This is consistent with our claims in~\cref{sec:from_prunning_to_mixed_precision}. In \cref{fig:sparse_win_selection}, the windows selection of the SparseViT model on the OOD sample is visualized. It can be observed that compared to \cref{fig:window-selection}, the window selection of SparseViT is distracted largely by the fog. As a result, many important windows are wrongly pruned, causing the performance degradation we observe. In the worst case, SparseViT has an even higher 43.4\% degradation rate compared to MixA-Q (30.6\%).

% \begin{figure}[htbp]
%   \centering
%   \begin{subfigure}[t]{0.8\linewidth}
%    \includegraphics[width=\linewidth]{figs/4_win.png}
%     \caption{stage 0 window selection of pure activation pruning}
%   \label{fig:foggy-win}
%   \end{subfigure}
%   \vfill
%   \begin{subfigure}[t]{0.8\linewidth}
%     \includegraphics[width=\linewidth]{figs/4.png}
%     \caption{Foggy weather}
%     \label{fig:foggy}
%   \end{subfigure}
%   \caption{Example of a short caption, which should be centered.}
% \label{fig:sparse_win_selection}
% \end{figure}

\subsection{QAT Result on COCO Panoptic Segmentation} 
We further evaluate MixA-Q on the COCO panoptic segmentation dataset. We use Mask2former with Swin-Tiny backbone, and the QAT settings are the same as in Sec. 4.1. As shown in \cref{fig:panoptic}, the performance enhancement (as in Sec. 4.1.2) after SAQA can still be observed, and 1.25$\times$ speedup is achieved with $\leq$ 1\% PQ drop. Compared to the object detection task, the relative performance of MixA-Q versus SparseViT improves on the segmentation task. MixA-Q surpasses SparseViT by large at relative cost $\geq$ 0.8. We speculate that this is because panoptic segmentation requires classifying every pixel (80 ``things'' and 53 ``stuff''), worsening performance degradation caused by pruning.

\subsection{PTQ Result on COCO} 

\paragraph{Setup}
We integrate MixA-Q with a representative PTQ method, RepQ~\cite{li2023repqvitscalereparameterizationposttraining}. As in QAT, we use Mask R-CNN with Swin-Tiny backbone and the COCO dataset for evaluation. The baseline model is quantized to W4A8. One random sample from the training set is used for calibration. When applying MixA-Q, the less important windows are assigned a bit width of 4. 

% The original RepQ method reparameterizes the weights and bias of the QKV layer inside MHA and the first layer of FFN. The updates are dependent on the activations channel-wise distributions, which conflicts with our sharing weights setup because the channel-wise distribution of high and low precision windows are not identical. To avoid this, we use the whole feature map as calibration data to initialize the step sizes of both high- and low-precision branches. 

RepQ makes post-training updates to the weights and bias of of the QKV layer and the first layer of FFN. To be compatible with our sharing weights setup, we use the full feature map as calibration data to initialize the step sizes of high- and low-precision branches so that the updates to the two branches' weights will be the same. As for the bias, we make them independent in PTQ. Note that during the evolutionary search, we evaluate different compression ratios without recalibrating the model.

% \vspace{-2mm}
% \paragraph{Result}
% \begin{figure}[htbp]
%     \centering
%     \includegraphics[width=0.5\linewidth]{figs/pareto_ptq_ev_w4a8_both.pdf}
%     \caption{Pareto front of MixA-Q and RepQ on two different calibration sets. The x-axis is the relative computation cost in terms of BOPs to the baseline model (W4A8 RepQ model). MixA-Q constantly has better computation cost and performance trade-off compared to the uniform-bit RepQ. Additionally, compared to RepQ on arbitrary bit widths (5, 6, 7 bits), our method uses a mixed 4- and 8-bit precision.}
% \label{fig:pareto-ptq}
% \vspace{-2mm}
% \end{figure}

\begin{figure}[!]
  \centering
  \begin{subfigure}[t]{0.49\linewidth}
    \includegraphics[width=\linewidth]{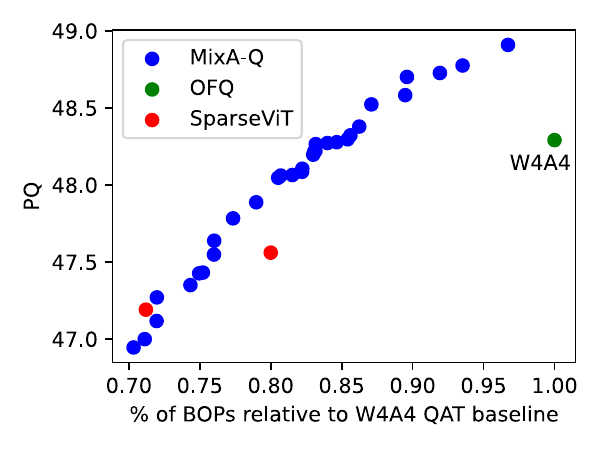}
    \caption{QAT Pareto front of MixA-Q and SparseViT on COCO Panoptic Segmentation dataset. Compared to the object detection task, the relative performance of MixA-Q versus SparseViT improves on the segmentation task.}
    \label{fig:panoptic}
  \end{subfigure}
  \hfill
  \begin{subfigure}[t]{0.49\linewidth}
    \includegraphics[width=\linewidth]{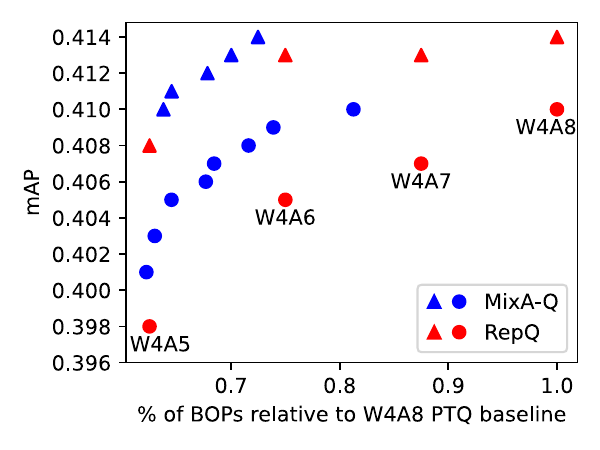}
    \caption{PTQ Pareto front of MixA-Q and RepQ on two different calibration sets. MixA-Q consistently has better computation cost and performance trade-off compared to the uniform-bit RepQ.}
    \label{fig:pareto-ptq}
  \end{subfigure}
  \caption{Further evaluation result on panoptic segmentation task and post-training quantization.}
  \vspace{-2mm}
\end{figure}
% \vspace{-12pt}
In \cref{fig:pareto-ptq}, we compare applying MixA-Q on the baseline model with uniformly reducing the activation bit widths of it from 8 to 5. We show the result on two different random calibration sets. It can be observed that MixA-Q constantly has better computational cost and performance trade-off compared to RepQ. Another benefit brought by MixA-Q is that bit widths from 5 to 7 are not supported by most of the general-purpose computing hardware, while our solution is based on a mixed 4-bit and 8-bit precision.

% \subsection{PTQ Result on COCO Panoptic}
% \paragraph{Setup}
% \paragraph{Result}
% \paragraph{Fine-tunig Result}

\subsection{Ablation Study}
\paragraph{SAQA from scratch}
In our experiments in \cref{sec:qat_coco}, we quantize the model into W4A4 before adding the low-precision branch and starting the SAQA. In this section, we explore whether directly adding a low-precision branch to the floating-point model and starting the SAQA has impact.

\begin{table}[t]
\centering
\begin{tabular}{l|ll}
\hline
                                   & Act bit & mAP  \\ \hline
\multirow{2}{*}{SAQA}              & 3.36*   & 43.4 \\
                                   & 3*      & 42.4 \\ \hline
\multirow{2}{*}{SAQA from fp}      & 3.36*   & 42.8 \\
                                   & 3*      & 42.1 \\ \hline
\end{tabular}
\caption{mAP comparison of SAQA based on model already quantized and SAQA from floating-point model}
\label{tab:saqa_ablation}
\vspace{-2mm}
\end{table}

From \cref{tab:saqa_ablation}, it can be observed that starting SAQA directly from the floating-point model results in worse mAP compared to starting SAQA based on the model that has been quantized on W4A4. Given that the loss landscape of ViT quantization is highly non-smooth~\cite{frumkin2024jumpinglocalminimaquantization}, we speculate that jointly training the two aggressively quantized branches makes the model converge sub-optimally.\\
\textbf{Windows Elevation:}
In MixA-Q, since less important windows are not pruned but instead kept in lower precision, it may make sense to elevate some windows to high precision again in later blocks. However, according to our experiments, allowing the elevation of previously compressed activation gives slightly worse (-0.1\%) or the same mAP. We speculate that the consistent computation precision of windows within one stage is important.

\section{Conclusion}

In this work, we introduced MixA-Q, a mixed-precision activation quantization method that revisits activation sparsity from a mixed-precision perspective to improve efficiency in window-based vision transformers. 
% Unlike prior mixed-precision methods that focus solely on inter-layer bit-width allocation, 
MixA-Q dynamically assigns high and low bit widts to different windows, effectively leveraging intra-layer activation sparsity. Compared to activation pruning methods that discard computations on less important windows, MixA-Q retains them in lower precision. This enables MixA-Q to be integrated training-free and gives greater robustness against out-of-distribution (OOD) inputs. 
% Additionally, MixA-Q can be combined with activation pruning and inter-layer mixed-precision methods, making it a versatile enhancement for model compression strategies.
On the COCO dataset, MixA-Q achieves a $1.35\times$ speedup in W4A8 PTQ settings and a $1.25\times$ speedup in W4A4 QAT settings without accuracy loss. By incorporating activation pruning, MixA-Q further achieves 1.5$\times$ with a 1\% mAP drop. 
% Additionally, through SAQA, MixA-Q can reduce the model's mAP by 24\% in W4A4 with a dynamic distillation effect. We hope our method can shed some light on the exploitation of activation sparsity with mixed-precision quantization.

\newpage
{
    \small
    \bibliographystyle{ieeenat_fullname}
    \bibliography{main}

\begin{thebibliography}{34}
\providecommand{\natexlab}[1]{#1}
\providecommand{\url}[1]{\texttt{#1}}
\expandafter\ifx\csname urlstyle\endcsname\relax
  \providecommand{\doi}[1]{doi: #1}\else
  \providecommand{\doi}{doi: \begingroup \urlstyle{rm}\Url}\fi

\bibitem[Blank and Deb(2020)]{9078759}
Julian Blank and Kalyanmoy Deb.
\newblock Pymoo: Multi-objective optimization in python.
\newblock \emph{IEEE Access}, 8:\penalty0 89497--89509, 2020.

\bibitem[Chen et~al.(2023)Chen, Liu, Tang, Yi, Zhao, and Han]{chen2023sparsevitrevisitingactivationsparsity}
Xuanyao Chen, Zhijian Liu, Haotian Tang, Li Yi, Hang Zhao, and Song Han.
\newblock Sparsevit: Revisiting activation sparsity for efficient high-resolution vision transformer.
\newblock In \emph{Proceedings of the IEEE/CVF conference on computer vision and pattern recognition}, pages 2061--2070, 2023.

\bibitem[Cheng et~al.(2022)Cheng, Misra, Schwing, Kirillov, and Girdhar]{cheng2022maskedattentionmasktransformeruniversal}
Bowen Cheng, Ishan Misra, Alexander~G Schwing, Alexander Kirillov, and Rohit Girdhar.
\newblock Masked-attention mask transformer for universal image segmentation.
\newblock In \emph{Proceedings of the IEEE/CVF conference on computer vision and pattern recognition}, pages 1290--1299, 2022.

\bibitem[Deb et~al.(2002)Deb, Pratap, Agarwal, and Meyarivan]{deb2002fast}
Kalyanmoy Deb, Amrit Pratap, Sameer Agarwal, and TAMT Meyarivan.
\newblock A fast and elitist multiobjective genetic algorithm: Nsga-ii.
\newblock \emph{IEEE transactions on evolutionary computation}, 6\penalty0 (2):\penalty0 182--197, 2002.

\bibitem[Dong et~al.(2023)Dong, Li, Wei, Niu, Tian, and Pan]{dong2023emqevolvingtrainingfreeproxies}
Peijie Dong, Lujun Li, Zimian Wei, Xin Niu, Zhiliang Tian, and Hengyue Pan.
\newblock Emq: Evolving training-free proxies for automated mixed precision quantization.
\newblock In \emph{Proceedings of the IEEE/CVF international conference on computer vision}, pages 17076--17086, 2023.

\bibitem[Dong et~al.(2019)Dong, Yao, Gholami, Mahoney, and Keutzer]{dong2019hawqhessianawarequantization}
Zhen Dong, Zhewei Yao, Amir Gholami, Michael~W Mahoney, and Kurt Keutzer.
\newblock Hawq: Hessian aware quantization of neural networks with mixed-precision.
\newblock In \emph{Proceedings of the IEEE/CVF international conference on computer vision}, pages 293--302, 2019.

\bibitem[Dong et~al.(2020)Dong, Yao, Arfeen, Gholami, Mahoney, and Keutzer]{dong2019hawqv2hessianawaretraceweighted}
Zhen Dong, Zhewei Yao, Daiyaan Arfeen, Amir Gholami, Michael~W Mahoney, and Kurt Keutzer.
\newblock Hawq-v2: Hessian aware trace-weighted quantization of neural networks.
\newblock \emph{Advances in neural information processing systems}, 33:\penalty0 18518--18529, 2020.

\bibitem[Dosovitskiy et~al.(2020)Dosovitskiy, Beyer, Kolesnikov, Weissenborn, Zhai, Unterthiner, Dehghani, Minderer, Heigold, Gelly, et~al.]{dosovitskiy2021imageworth16x16words}
Alexey Dosovitskiy, Lucas Beyer, Alexander Kolesnikov, Dirk Weissenborn, Xiaohua Zhai, Thomas Unterthiner, Mostafa Dehghani, Matthias Minderer, Georg Heigold, Sylvain Gelly, et~al.
\newblock An image is worth 16x16 words: Transformers for image recognition at scale.
\newblock \emph{arXiv preprint arXiv:2010.11929}, 2020.

\bibitem[Fayyaz et~al.(2022)Fayyaz, Koohpayegani, Jafari, Sengupta, Joze, Sommerlade, Pirsiavash, and Gall]{fayyaz2022adaptivetokensamplingefficient}
Mohsen Fayyaz, Soroush~Abbasi Koohpayegani, Farnoush~Rezaei Jafari, Sunando Sengupta, Hamid Reza~Vaezi Joze, Eric Sommerlade, Hamed Pirsiavash, and J{\"u}rgen Gall.
\newblock Adaptive token sampling for efficient vision transformers.
\newblock In \emph{European Conference on Computer Vision}, pages 396--414. Springer, 2022.

\bibitem[Frumkin et~al.(2023)Frumkin, Gope, and Marculescu]{frumkin2024jumpinglocalminimaquantization}
Natalia Frumkin, Dibakar Gope, and Diana Marculescu.
\newblock Jumping through local minima: Quantization in the loss landscape of vision transformers.
\newblock In \emph{Proceedings of the IEEE/CVF International Conference on Computer Vision}, pages 16978--16988, 2023.

\bibitem[He et~al.(2017)He, Gkioxari, Doll{\'a}r, and Girshick]{he2018maskrcnn}
Kaiming He, Georgia Gkioxari, Piotr Doll{\'a}r, and Ross Girshick.
\newblock Mask r-cnn.
\newblock In \emph{Proceedings of the IEEE international conference on computer vision}, pages 2961--2969, 2017.

\bibitem[Kim et~al.(2024)Kim, Lee, Yoo, and Kim]{Kim_2024}
Han-Byul Kim, Joo~Hyung Lee, Sungjoo Yoo, and Hong-Seok Kim.
\newblock Metamix: Meta-state precision searcher for mixed-precision activation quantization.
\newblock \emph{Proceedings of the AAAI Conference on Artificial Intelligence}, 38\penalty0 (12):\penalty0 13132–13141, 2024.

\bibitem[Kong et~al.(2022)Kong, Dong, Ma, Meng, Niu, Sun, Shen, Yuan, Ren, Tang, et~al.]{kong2022spvitenablingfastervision}
Zhenglun Kong, Peiyan Dong, Xiaolong Ma, Xin Meng, Wei Niu, Mengshu Sun, Xuan Shen, Geng Yuan, Bin Ren, Hao Tang, et~al.
\newblock Spvit: Enabling faster vision transformers via latency-aware soft token pruning.
\newblock In \emph{European conference on computer vision}, pages 620--640. Springer, 2022.

\bibitem[Li et~al.(2022)Li, Xu, Zhang, Cao, Gao, and Guo]{li2022qvitaccuratefullyquantized}
Yanjing Li, Sheng Xu, Baochang Zhang, Xianbin Cao, Peng Gao, and Guodong Guo.
\newblock Q-vit: Accurate and fully quantized low-bit vision transformer.
\newblock \emph{Advances in neural information processing systems}, 35:\penalty0 34451--34463, 2022.

\bibitem[Li et~al.(2023)Li, Xiao, Yang, and Gu]{li2023repqvitscalereparameterizationposttraining}
Zhikai Li, Junrui Xiao, Lianwei Yang, and Qingyi Gu.
\newblock Repq-vit: Scale reparameterization for post-training quantization of vision transformers.
\newblock In \emph{Proceedings of the IEEE/CVF International Conference on Computer Vision}, pages 17227--17236, 2023.

\bibitem[Liang et~al.(2022)Liang, Ge, Tong, Song, Wang, and Xie]{liang2022patchesneedexpeditingvision}
Youwei Liang, Chongjian Ge, Zhan Tong, Yibing Song, Jue Wang, and Pengtao Xie.
\newblock Not all patches are what you need: Expediting vision transformers via token reorganizations.
\newblock \emph{arXiv preprint arXiv:2202.07800}, 2022.

\bibitem[Lin et~al.(2014)Lin, Maire, Belongie, Hays, Perona, Ramanan, Doll{\'a}r, and Zitnick]{lin2015microsoftcococommonobjects}
Tsung-Yi Lin, Michael Maire, Serge Belongie, James Hays, Pietro Perona, Deva Ramanan, Piotr Doll{\'a}r, and C~Lawrence Zitnick.
\newblock Microsoft coco: Common objects in context.
\newblock In \emph{Computer vision--ECCV 2014: 13th European conference, zurich, Switzerland, September 6-12, 2014, proceedings, part v 13}, pages 740--755. Springer, 2014.

\bibitem[Liu et~al.(2024{\natexlab{a}})Liu, Gong, Wei, Dong, Cai, and Zhuang]{liu2024qllmaccurateefficientlowbitwidth}
Jing Liu, Ruihao Gong, Xiuying Wei, Zhiwei Dong, Jianfei Cai, and Bohan Zhuang.
\newblock {QLLM}: Accurate and efficient low-bitwidth quantization for large language models.
\newblock In \emph{International Conference on Learning Representations (ICLR)}, 2024{\natexlab{a}}.

\bibitem[Liu et~al.(2023)Liu, Liu, and Cheng]{liu2023oscillationfreequantizationlowbitvision}
Shih-Yang Liu, Zechun Liu, and Kwang-Ting Cheng.
\newblock Oscillation-free quantization for low-bit vision transformers.
\newblock In \emph{International conference on machine learning}, pages 21813--21824. PMLR, 2023.

\bibitem[Liu et~al.(2024{\natexlab{b}})Liu, Ding, Yu, Xi, Li, Tu, Hu, Chen, Yin, and Xiong]{liu2024pq}
Xiaoyu Liu, Xin Ding, Lei Yu, Yuanyuan Xi, Wei Li, Zhijun Tu, Jie Hu, Hanting Chen, Baoqun Yin, and Zhiwei Xiong.
\newblock Pq-sam: Post-training quantization for segment anything model.
\newblock In \emph{European Conference on Computer Vision}, pages 420--437. Springer, 2024{\natexlab{b}}.

\bibitem[Liu et~al.(2024{\natexlab{c}})Liu, Gehrig, Messikommer, Cannici, and Scaramuzza]{liu2023revisitingtokenpruningobject}
Yifei Liu, Mathias Gehrig, Nico Messikommer, Marco Cannici, and Davide Scaramuzza.
\newblock Revisiting token pruning for object detection and instance segmentation.
\newblock In \emph{Proceedings of the IEEE/CVF Winter Conference on Applications of Computer Vision}, pages 2658--2668, 2024{\natexlab{c}}.

\bibitem[Liu et~al.(2021)Liu, Lin, Cao, Hu, Wei, Zhang, Lin, and Guo]{liu2021swintransformerhierarchicalvision}
Ze Liu, Yutong Lin, Yue Cao, Han Hu, Yixuan Wei, Zheng Zhang, Stephen Lin, and Baining Guo.
\newblock Swin transformer: Hierarchical vision transformer using shifted windows.
\newblock In \emph{Proceedings of the IEEE/CVF international conference on computer vision}, pages 10012--10022, 2021.

\bibitem[Ma et~al.(2023)Ma, Jin, Zheng, Wang, Li, Wu, Jiang, Zhang, and Ji]{ma2022ompqorthogonalmixedprecision}
Yuexiao Ma, Taisong Jin, Xiawu Zheng, Yan Wang, Huixia Li, Yongjian Wu, Guannan Jiang, Wei Zhang, and Rongrong Ji.
\newblock Ompq: Orthogonal mixed precision quantization.
\newblock In \emph{Proceedings of the AAAI conference on artificial intelligence}, pages 9029--9037, 2023.

\bibitem[Mao et~al.(2023)Mao, Chen, Zhu, Chen, Su, Zhang, and Xue]{mao2023cocoobenchmarkobjectdetectors}
Xiaofeng Mao, Yuefeng Chen, Yao Zhu, Da Chen, Hang Su, Rong Zhang, and Hui Xue.
\newblock Coco-o: A benchmark for object detectors under natural distribution shifts.
\newblock In \emph{Proceedings of the IEEE/CVF International Conference on Computer Vision}, pages 6339--6350, 2023.

\bibitem[Nagel et~al.(2019)Nagel, Baalen, Blankevoort, and Welling]{nagel2019datafreequantizationweightequalization}
Markus Nagel, Mart~van Baalen, Tijmen Blankevoort, and Max Welling.
\newblock Data-free quantization through weight equalization and bias correction.
\newblock In \emph{Proceedings of the IEEE/CVF international conference on computer vision}, pages 1325--1334, 2019.

\bibitem[Nagel et~al.(2020)Nagel, Amjad, Van~Baalen, Louizos, and Blankevoort]{nagel2020downadaptiveroundingposttraining}
Markus Nagel, Rana~Ali Amjad, Mart Van~Baalen, Christos Louizos, and Tijmen Blankevoort.
\newblock Up or down? adaptive rounding for post-training quantization.
\newblock In \emph{International conference on machine learning}, pages 7197--7206. PMLR, 2020.

\bibitem[Rao et~al.(2021)Rao, Zhao, Liu, Lu, Zhou, and Hsieh]{rao2021dynamicvitefficientvisiontransformers}
Yongming Rao, Wenliang Zhao, Benlin Liu, Jiwen Lu, Jie Zhou, and Cho-Jui Hsieh.
\newblock Dynamicvit: Efficient vision transformers with dynamic token sparsification.
\newblock \emph{Advances in neural information processing systems}, 34:\penalty0 13937--13949, 2021.

\bibitem[Sun et~al.(2022)Sun, Ge, Wang, Lin, Chen, Li, and Sun]{sun2022entropy}
Zhenhong Sun, Ce Ge, Junyan Wang, Ming Lin, Hesen Chen, Hao Li, and Xiuyu Sun.
\newblock Entropy-driven mixed-precision quantization for deep network design.
\newblock \emph{Advances in Neural Information Processing Systems}, 35:\penalty0 21508--21520, 2022.

\bibitem[Vaswani et~al.(2017)Vaswani, Shazeer, Parmar, Uszkoreit, Jones, Gomez, Kaiser, and Polosukhin]{vaswani2023attentionneed}
Ashish Vaswani, Noam Shazeer, Niki Parmar, Jakob Uszkoreit, Llion Jones, Aidan~N Gomez, {\L}ukasz Kaiser, and Illia Polosukhin.
\newblock Attention is all you need.
\newblock \emph{Advances in neural information processing systems}, 30, 2017.

\bibitem[Wang et~al.(2024)Wang, Zhang, Li, Xu, Miao, and Wang]{wang2024thinkinggranularitydynamicquantization}
Mingshen Wang, Zhao Zhang, Feng Li, Ke Xu, Kang Miao, and Meng Wang.
\newblock Thinking in granularity: Dynamic quantization for image super-resolution by intriguing multi-granularity clues, 2024.

\bibitem[Wang et~al.(2020)Wang, Wang, Cai, Lin, Liu, Wang, Lin, and Han]{wang2020apqjointsearchnetwork}
Tianzhe Wang, Kuan Wang, Han Cai, Ji Lin, Zhijian Liu, Hanrui Wang, Yujun Lin, and Song Han.
\newblock Apq: Joint search for network architecture, pruning and quantization policy.
\newblock In \emph{Proceedings of the IEEE/CVF Conference on Computer Vision and Pattern Recognition}, pages 2078--2087, 2020.

\bibitem[Xiao et~al.(2023)Xiao, Li, Yang, and Gu]{xiao2023patchwisemixedprecisionquantizationvision}
Junrui Xiao, Zhikai Li, Lianwei Yang, and Qingyi Gu.
\newblock Patch-wise mixed-precision quantization of vision transformer, 2023.

\bibitem[Zhang et~al.(2021)Zhang, Pang, Chen, and Loy]{zhang2021knetunifiedimagesegmentation}
Wenwei Zhang, Jiangmiao Pang, Kai Chen, and Chen~Change Loy.
\newblock K-net: Towards unified image segmentation.
\newblock \emph{Advances in Neural Information Processing Systems}, 34:\penalty0 10326--10338, 2021.

\bibitem[Zhong et~al.(2025)Zhong, Huang, Hu, Zhang, and Ji]{zhong2025accurateposttrainingquantizationvision}
Yunshan Zhong, You Huang, Jiawei Hu, Yuxin Zhang, and Rongrong Ji.
\newblock Towards accurate post-training quantization of vision transformers via error reduction.
\newblock \emph{IEEE Transactions on Pattern Analysis and Machine Intelligence}, 2025.

\end{thebibliography}
}

\newpage
%-------------------------------------------------------------------------
\clearpage
\section{Appendix}
\subsection{Uniform-sum Compression Ratio Sampling}
We share the details of our sampling algorithm here:

\begin{algorithm}[h]
\caption{Uniform-Sum Compression Ratio Sampling}
\begin{algorithmic}[1]
\State \textbf{Input:} Number of layers $n$, predefined sum range $[s_{\min}, s_{\max}]$, upper bound $u$
\State \textbf{Output:} compression ratios $\{r_0, r_1, \dots, r_n\}$ such that $\sum r_i = S$

\State  $S_{\min} \gets s_{\min} \times 10$, $S_{\max} \gets s_{\max} \times 10$, $U \gets u \times 10$

\State Sample a target sum of compression ratios:
\State \hspace{1cm} $S \sim \text{Uniform}(S_{\min}, S_{\max})$

\Repeat
    \State Sample compression ratios from a Dirichlet distribution:
    \State \hspace{1cm} $(r_0, r_1, \dots, r_n) \sim \text{Dirichlet}(\alpha)$
    \State Scale sampled values to ensure their sum equals $S$:
    \State \hspace{1cm} $r_i \gets r_i \cdot S$, for all $i$
\Until{$r_i \leq U$ for all $i$}

\State \textbf{Apply Largest Remainder Method (LRM) for rounding:}
\State Round down each element: $R \gets \lfloor r_i \rfloor$
\State Compute remaining difference: $\text{diff} \gets S - \sum R$

\If{$\text{diff} > 0$}
    \State Compute fractional remainders: $\text{remainder} \gets r_i - R$
    \State Sort indices by largest remainder in descending order
    \For{$i = 1$ to $\text{diff}$}
        \State Increment $R$ at the index with the highest remainder
    \EndFor
\EndIf

\For{$i = 0$ to $n$}
    \State $r_i \gets R_i / 10$
\EndFor

\State \textbf{Return} $\{r_0, r_1, \dots, r_n\}$
\end{algorithmic}
\end{algorithm}

For the Swin-Tiny backbone that we use in the experiments, the number of blocks is 6, and the upper bound for compression ratios is 0.8.

We aim for the relative computational cost to fall within the range $[0.65,0.95]$. Given that MixA-Q reduces computations by 50\%, the corresponding range for the sum of compression ratios is computed as:

\(
[(1-0.95)\times6\times2,(1-0.65)\times6\times2] = [0.6,4.2]
\)

\noindent Note that during sampling, we assume that all blocks have an equal number of BOPs. However, in our experiments, the reported relative computational cost is recalculated based on the actual distribution of BOPs across different blocks. 

% Please add the following required packages to your document preamble:
% \usepackage{multirow}
\begin{table}[h]
\centering
\begin{tabular}{l|ll}
\hline
                               & Act Bit & mAP  \\ \hline
\multirow{3}{*}{SAQA  uni.sum} & 3.36*   & 43.4 \\
                               & 3.23*   & 43.2 \\
                               & 3*      & 42.4 \\ \hline
SAQA                           & 3.36*   & 43.0 \\
                               & 3.23*   & 42.8 \\
                               & 3*      & 41.9 \\ \hline
\end{tabular}
\caption{mAP performance comparison of SAQA using uniform-sum sampling and naive uniform sampling. * indicates that the bit width is a computationally weighted average.}
\label{tab:saqa_compare}
\end{table}

In \cref{tab:saqa_compare} we compare the SAQA using uniform-sum sampling and naive uniform sampling. It can be observed that SAQA with uniform-sum sampling constantly achieves higher mAP at different averaged activation bits. 

\begin{figure}[htbp]
  \centering
    \includegraphics[width=0.8\linewidth]{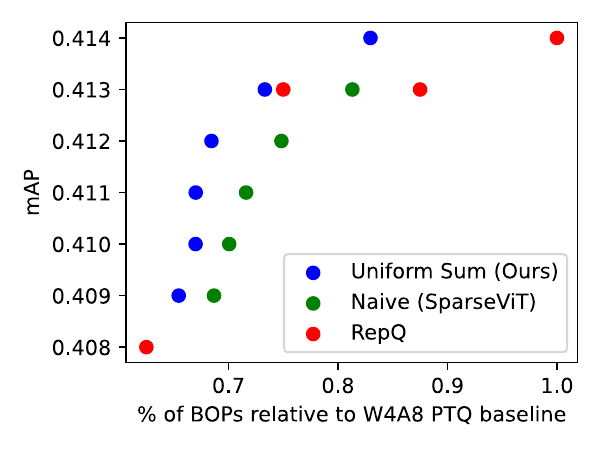}
    \caption{Pareto front after two generations of evolutionary search using different sampling methods.}
    \label{fig:es}
\end{figure}

In terms of sampling efficiency, as shown in \cref{fig:es}, our uniform-sum sampling is able to find configurations for MixA-Q that is better than the PTQ baseline after only two generations of the evolutionary search, while naive sampling can’t. This means that our uniform-sum sampling helps the search process to converge faster and thus saves searching time.

\subsection{Calculation of Equivalent Activation Bits}

In this section we explain how the equivalent activation bits (marked with *) in our experiment part are calculated for MixA-Q and SparseViT. Given the GMACs (Giga Multiply-Accumulate operations) of the Swin-Tiny backbone:

\begin{align}
    &\text{GMACs} = [13.3, 13.56, 14.16, 14.16, 14.16, 14.02] \\
    &C = 3
\end{align}

GMACs contains the number of GMACs for each pair of consecutive Swin blocks and the C is the number of GMACs of for downsampling layers whose computations can't be saved by MixA-Q or SparseViT.

In MixA-Q, 50\% computations of the compressed windows are saved while 100\% computations of the pruned windows are saved in SparseViT. Thus, for a given compression or pruning ratio configuration \( RC \) and the activation bit of the baseline model ActBit$_{base}$ the equivalent activation bits ActBit$_{eq}$ is calculated as:

\begin{align}
    \text{ratios} &= \left( 1 - \frac{r}{2} \right), \quad \forall r \in RC, \text{if method is MixA-Q} \\
    \text{ratios} &= 1-r, \quad \forall r \in RC, \text{if method is SparseViT} \\
    \text{ActBit}_{eq} &= \text{ActBit}_{base} \times \frac{\sum_{i} \left( \text{ratios}_i \cdot \text{GMACs}_i \right) + C}
         {\sum_{i} \text{GMACs}_i + C}
\end{align}

\end{document}